\pdfoutput = 1
\documentclass[twocolumn]{article}

\usepackage{times}
\usepackage[english]{babel}
\usepackage{blindtext}
\usepackage{graphicx}
\usepackage{amsmath, amsthm, amssymb, bbm, bm}
\usepackage{float}      
\usepackage{subcaption}  
\usepackage{wrapfig}
\usepackage[margin=0cm]{caption}
\usepackage[titletoc, toc]{appendix}
\usepackage{tabularx}

\usepackage{multirow}
\usepackage{hhline}
\usepackage{makecell}
\usepackage{placeins}  

\usepackage[x11names, usenames, dvipsnames, svgnames, table]{xcolor}
\definecolor{firebrick}{rgb}{.698,.133,.133}
\definecolor{mybluelight}{rgb}{0.9, 0.9, 1.}
\definecolor{myorangelight}{rgb}{1., 0.9, 0.9}

\usepackage[utf8]{inputenc} 
\usepackage[T1]{fontenc}    
\usepackage{url}            
\usepackage{booktabs, colortbl}       
\usepackage{amsfonts}       
\usepackage{nicefrac}       
\usepackage{microtype}      

\usepackage{csquotes}
\usepackage{latexsym}

\usepackage{pifont}
\usepackage[boxruled, vlined, linesnumbered]{algorithm2e}
\SetAlFnt{\small}
\SetAlCapFnt{\small}
\SetAlCapNameFnt{\small}
\usepackage{algorithmic}
\algsetup{linenosize=\tiny}

\let\oldnl\nl
\newcommand{\nonl}{\renewcommand{\nl}{\let\nl\oldnl}}

\usepackage{paralist}

\usepackage{xspace}
\usepackage{soul}
\usepackage{dsfont}
\usepackage{stmaryrd}
\usepackage[textwidth=15mm]{todonotes}
\usepackage{dirtytalk}
\usepackage{pbox}
\usepackage{cprotect}

\usepackage{verbatim}
\usepackage{textcomp}
\usepackage[normalem]{ulem}

\usepackage{mathtools}
\usepackage{etextools}

\usepackage[colorlinks=true,allcolors=firebrick,bookmarks=false]{hyperref}

\usepackage{graphicx}
\usepackage{amsmath}
\usepackage{amssymb}
\newsavebox{\foobox}
\newcommand{\slantbox}[2][.3]
  {%
    \mbox
      {%
        \sbox{\foobox}{#2}%
        \hskip\wd\foobox
        \pdfsave
        \pdfsetmatrix{1 0 #1 1}%
        \llap{\usebox{\foobox}}%
        \pdfrestore
      }%
  }

\usepackage{booktabs}
\usepackage{colortbl}

\usepackage{xcolor}
\usepackage{enumerate}

\newcommand\maxboxacc{\texttt{MaxBoxAcc}\xspace}
\newcommand\newmaxboxacc{\texttt{MaxBoxAccV2}\xspace}

\newcommand\cls{\texttt{class}\xspace}
\newcommand\topone{\texttt{top-1}\xspace}
\newcommand\topfive{\texttt{top-5}\xspace}
\newcommand\ltopone{\texttt{Top-1}\xspace}
\newcommand\ltopfive{\texttt{Top-5}\xspace}

\newcommand\ioa{\texttt{IoA}\xspace}
\newcommand\iop{\texttt{IoP}\xspace}

\newcommand\iou{\texttt{IoU}\xspace}

\definecolor{darkergreen}{RGB}{21, 152, 56}
\definecolor{red2}{RGB}{252, 54, 65}
\definecolor{Gray}{gray}{0.85}
\newcolumntype{g}{>{\columncolor{Gray}}c}

\definecolor{darkergreen}{RGB}{21, 152, 56}
\definecolor{red2}{RGB}{252, 54, 65}
\definecolor{Gray}{gray}{0.85}
\newcolumntype{g}{>{\columncolor{Gray}}c}

\theoremstyle{definition}

\usepackage{stfloats}
\DeclarePairedDelimiterX{\divx}[2]{(}{)}{%
  #1\;\delimsize\|\;#2%
}

\newcommand*{\ie}{\emph{i.e.}\@\xspace}

\usepackage{style}

\makeatletter
\@namedef{ver@everyshi.sty}{}
\newcommand{\removelatexerror}{\let\@latex@error\@gobble}

\title{Discriminative Sampling of Proposals in Self-Supervised Transformers\\ for Weakly Supervised Object Localization}

\renewcommand\footnotemark{}

\author{Shakeeb~Murtaza$^{1}$,
  ~Soufiane~Belharbi$^{1}$,
  ~Marco~Pedersoli$^{1}$,
  ~Aydin~Sarraf$^{2}$, and
  ~Eric~Granger$^{1}$\\
 $^1$ LIVIA, Dept. of Systems Engineering, ETS Montreal, Canada \\
$^2$  Ericsson, Global AI Accelerator, Montreal, Canada\\
{\tt\footnotesize \textcolor{black}{shakeeb.murtaza.1@ens.etsmtl.ca} }
}

\newcommand{\ignore}[1]{}

\newcommand\lpe{\texttt{LPE}\xspace}
\newcommand\lme{\texttt{LME}\xspace}

 \usepackage[shortlabels]{enumitem}

\begin{document}
\maketitle\thispagestyle{fancy}

\maketitle
\lhead{\color{gray} \small \today}
\rhead{\color{gray} \small Murtaza et al. \;  [WACV Workshop 2023]}

\begin{abstract}
Drones are employed in a growing number of visual recognition applications. A recent development in cell tower inspection is drone-based asset surveillance, where the autonomous flight of a drone is guided by localizing objects of interest in successive aerial images. In this paper, we propose a method to train deep weakly-supervised object localization (WSOL) models  based only on image-class labels to locate object with high confidence. To train our localizer, pseudo labels are efficiently harvested from a self-supervised vision transformers (SSTs). However, since SSTs decompose the scene into multiple maps containing various object parts, and do not rely on any explicit supervisory signal, they cannot distinguish between the object of interest and other objects, as required WSOL. To address this issue, we propose leveraging the multiple maps generated by the different transformer heads to acquire pseudo-labels for training a deep WSOL model. In particular, a new \textbf{Di}scriminative \textbf{P}roposals \textbf{S}ampling (DiPS) method is introduced that relies on a CNN classifier to identify discriminative regions. Then, foreground and background pixels are sampled from these regions in order to train a WSOL model for generating activation maps that can accurately localize objects belonging to a specific class. Empirical results\footnote{Our code is available: 
\href{https://github.com/shakeebmurtaza/dips}{https://github.com/shakeebmurtaza/dips}}
on the challenging TelDrone dataset indicate that our proposed approach can outperform state-of-art methods over a wide range of threshold values over produced maps. We also computed results on CUB dataset, showing that our method can be adapted for other tasks.
\end{abstract}

\section{Introduction}
\label{sec:intro}

Due to its efficiency and flexibility, drone based surveillance has recently emerged as a feasible alternative for monitoring assets at numerous cell tower sites. Globally, millions of cell towers are being monitored for verification of assets, e.g., antennas. However, manual cell tower inspection is highly dangerous and expensive \cite{allred_2022}. To deploy drones for surveillance we need to localize objects that help drones to fly autonomously. Additionally, once a drone is able to identify and localize objects, it can perform different visual recognition tasks. For visual recognition, the drone should be able to fly at a safe distance from obstacles. Given the difficulty incurred in acquiring aerial images of concerned object at different viewpoints, and their associated bounding box annotations, we are unable to employ object localization models  trained using supervised learning. To deal with this issue, we propose using model that can efficiently learn to localize objects of interest in a weakly-supervised manner \cite{choe2021region}, and efficiently guide the drone as it captures cell tower. 

Commonly used methods for weakly supervised object localization are based on class activation maps (CAMs) built on top of standard convolutional neural networks (CNNs)~\cite{belharbi2022fcam,choe2019attention,lee2019ficklenet,rahimi2020pairwise,singh2017hide,wei2021shallow,wei2017object,xue2019danet,yang2020combinational,yun2019cutmix,zhang2020rethinking,zhang2018self}. These methods allow producing a spatial class activation map highlighting features belonging to a particular class using features from the penultimate layer \cite{zhou2016learning} of a CNN. Strong activation of a map indicates the potential presence of the corresponding class, allowing for object localization. To improve the robustness of CAM methods, several techniques have been proposed to obtain maps from different layers by utilizing gradient information \cite{fu2020axiom, selvaraju2017grad, chattopadhay2018grad}. Despite their popularity, CAM methods have limitations leading to inaccurate localization. Activation maps tend to focus on small discriminant areas of objects \cite{belharbi2022fcam, belharbi2022tcam} -- common to different instances of an object belonging to a specific class. Current efforts in the WSOL litterature focus on improving the CAMs to cover the full object. However, for aerial cell tower photos, regions of interest (RoI) are quite small relative to the entire images, and can  be quite distant from the drone. Hence, current CAM-based methods produce bloby results, and are unable to adequately localize objects \cite{belharbi2022fcam, murtaza2022constrained}.
 
\begin{figure}[!t]
\begin{center}
\includegraphics[width=1\linewidth,trim=4 4 4 4, clip]{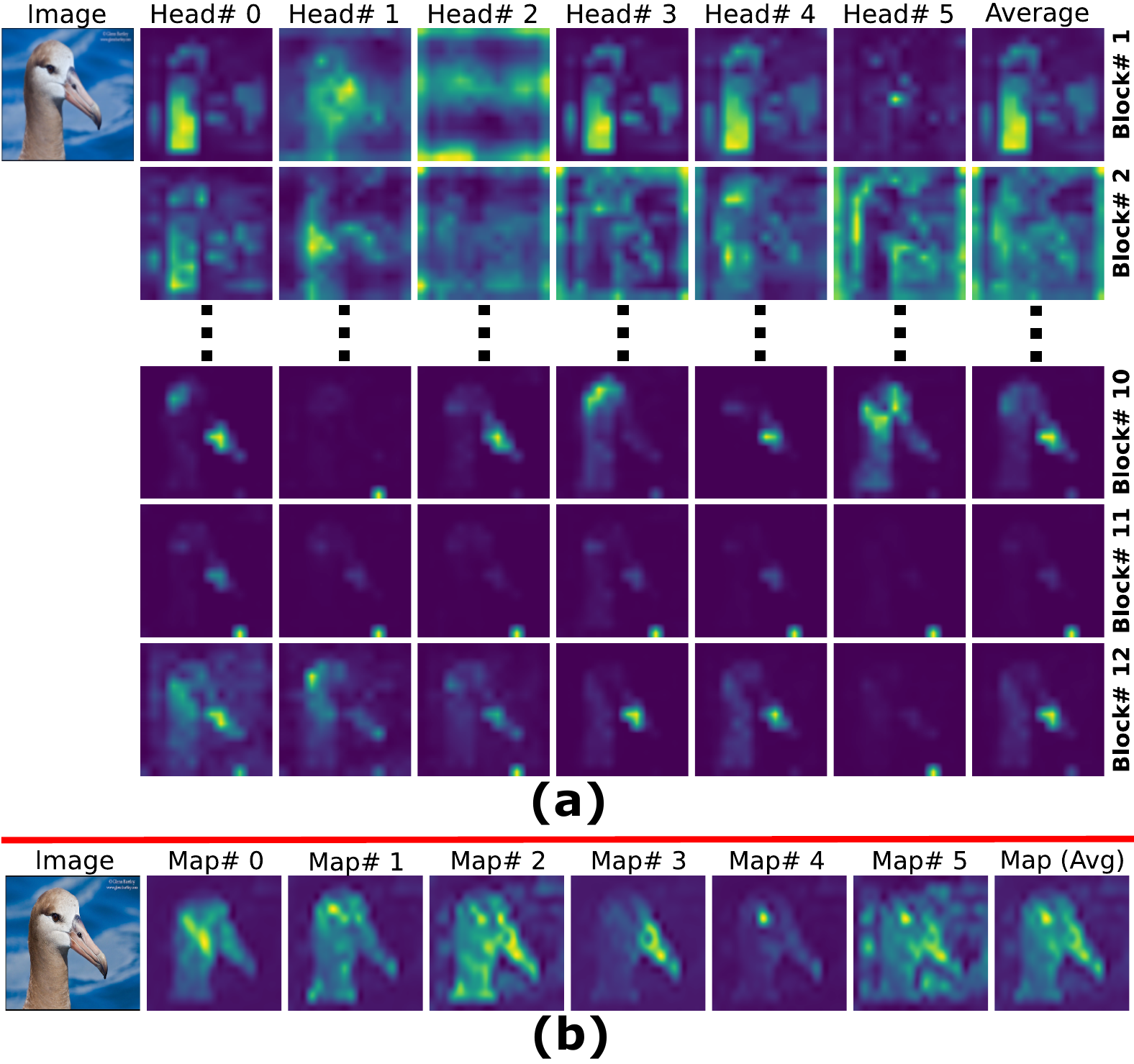}
\end{center}
\caption{\textbf{(a):} Attention of each \cls token learned by TS-CAM \cite{gao2021ts} map from the first two and last three blocks. The first half of the blocks attends to different features, including background regions, but TS-CAM is able to accumulate all of them. Attention maps of each \cls token are shown in Supplementary Material. \textbf{(b):} The attention map of \cls tokens corresponding to each token head from the last layer of transformer learned in a self-supervised fashion. These maps show that the last layer is capable of accumulating fine-grained localization of the concerned object.}\label{fig:dino_last_head}
\end{figure}

Recently, self-supervised transformers (SSTs)~\cite{caron2021emerging} has attracted much attention for WSOL tasks. Using only self-supervision, these models can yield impressive saliency maps. Given their long-range receptive fields, SSTs analyze an entire input image, allowing them to build saliency maps that cover full objects. Transformers are able to identify multiple objects by decomposing the scene into different spatial saliency maps. Such localization information is accumulated in maps, \ie \cls tokens, at the top layer (see Fig.~\ref{fig:dino_last_head}(b)). However, without any class supervision, these tokens are arbitrary and class-agnostic, and objects of interest cannot be dissociated from others. Each token focuses on a random object with semantics that differ from one image to another, making them less reliable for localization. Unless ground truth localizations are provided to select the best token~\cite{caron2021emerging}, their application in WSOL remains limited.

Few recent work have been proposed to extract discriminative localization from \cls tokens. In particular, TS-CAM~\cite{gao2021ts} has been proposed to leverage localization in transformers, where the average of all \cls tokens is multiplied with a semantic-aware map to produce the final activation map. Since TS-CAM averages an abundance of class-agnostic attention maps, it is prone to localization error by including non-discriminative regions. Additionally, maps from earlier layers attend to background regions, as shown in Fig.\ref{fig:dino_last_head}(a). Therefore, accumulating all maps introduces noise in the final activation map that can reduce the model's localization accuracy. Different transformer-based methods have been proposed \cite{bai2022weakly, chen2022lctr, gupta2022vitol, su2022re}. \cite{bai2022weakly} propose a spatial calibration module to capture semantic relationships and spatial similarities. Similarly, ViTOL \cite{gupta2022vitol} proposes to incorporate a patch-based attention dropout layer into the transformer attention blocks to improve localization maps. \cite{chen2022lctr} proposes a relational patch-attention module to enhance local perception quality, and to retain global information. \cite{su2022re} introduces a mechanism to suppress background regions and focus on the target object. All these methods are unable to minimize the loss over the localization maps as they solely focus on classification tasks, limiting their ability to fully cover an object while requiring a search for an optimal threshold.

In this paper, a new Discriminative Proposals Sampling (DiPS) method is proposed to leverage localization information from SSTs for accurate WSOL in aerial images captured by drones. During training, \cls token maps are extracted from the top block of pretrained SST. Given a limited number of high-quality attention maps (\cls tokens) covering different objects, we employ a CNN classifier to localize potential discriminative regions (see Fig. \ref{fig:dino_last_head}(a)). Then, pixel-wise pseudo-labels are sampled from these regions to train a U-Net style localization network. To benefit from these tokens, a method is introduced for collecting appropriate pseudo-labels, and sampling foreground/background regions from them to train our localization network such that a particular object can be localized with a high level of confidence (Fig.\ref{fig:our_method}). In particular, with the help of a CNN classifier, sampling areas are identified for building effective pseudo-labels by determining foreground and background regions in \cls tokens obtained from SST. From these areas, foreground pixels are sampled, while the entire image is used to sample pixels from background regions according to the criteria defined in Section \ref{sec:method}. Using these pseudo-pixels, the localization network is trained by optimizing partial cross-entropy over selected pixels. Additionally, the localization network parameters are regularized by the classifier response to ensure consistency between class prediction and localized area. CRF loss \cite{tang2018regularized} is employed to yield localization with accurate boundaries. During inference, we retain only the localization network to produce of localization maps, and the CNN classifier to predict the class. 

Our DiPS method allows producing localization maps with same size as the input image. Existing transformer and CNN-based methods can only produce low-resolution maps which introduce more localization error. For CNN-based methods, the class activation map is approximately $8 \times$ smaller than the input image, and requires interpolation that adds a bloby effect. In contrast to CNNs, which provide a map of size $29 \times 29$, the transformer-based methods can produce a map of size $14 \times 14$ (with standard architecture) for an input image of $224 \times 224$. Producing low-resolution maps hinders the performance of localization methods, and we cannot go beyond a threshold in terms of localization accuracy. These maps are unable to encompass precise details about the concerned object of interest. Also, other methods that use activation maps as hard pseudo labels are only able to focus on representative parts of an object which hinders their performance.  In contrast, our method selects few pixels as pseudo label that prevent the model from learning false-positives from underlying CAMs.

\noindent \textbf{Our main contributions are summarized as follows.}

\noindent \textbf{(1)} A new DiPS method is proposed to leverage the emerging saliency maps in SST to localize cell towers in aerial images. Since maps from SSTs   
lack discriminative information, a pretrained CNN classifier is employed to produce discriminative proposals based on saliency maps from the top layer of pretrained SST to harvest efficient pseudo-labels for the training of our localizer.

\noindent \textbf{(2)} Instead of minimizing the classification loss, DiPS samples the pseudo labels, allowing to minimize the loss over localization maps.

\noindent \textbf{(3)} DiPS can efficiently infer statistical properties (e.g., size, boundaries) of the object learned from low-resolution maps, resulting in a high-quality localization map.

\noindent \textbf{(4)} An extensive set of experiments was conducted on two challenging WSOL datasets -- (i) TelDrone, a private dataset containing aerial images from various cell tower sites, and (ii) CUB-200-2011 \cite{WelinderEtal2010}. The proposed method outperformed state-of-art methods in terms of localization accuracy, with less sensitivity to threshold values. Visual results also show that our method produces CAMs with a better coverage of the entire foreground regions, and a clearer distinction between foreground and background regions.

\section{Related Work}\label{sec:relatedwork}

\noindent\textbf{Class Activation Mapping (CAM) Methods:}
A common way to harvest localization maps from network activations is by aggregating the information presented in a specific CNN layer. The representative method for aggregating activations is using CAM methods~\cite{zhou2016learning}, which weighs each pixel according to its influence on the class prediction. Several methods have been proposed to improve the mechanism for harvesting the activation maps 
\cite{chattopadhay2018grad,fu2020axiom, ramaswamy2020ablation,selvaraju2017grad}. Networks are used for generating CAMs are solely trained for image classification, and thus focus on discriminative regions. They typically  underestimate the object size. To mitigate this issue, \cite{singh2017hide,yun2019cutmix} employ adversarial perturbation to erase discriminative parts, and for the network to look beyond the discriminative areas. Similarly, in \cite{choe2019attention,zhang2018adversarial} discriminative features are erased, and adversarial learning is adopted to encroach upon non-discriminative regions of the concerned object. In addition, fusion-based methods improve on CAM methods by combining different activation maps based on the classifier's response \cite{naidu2020cam, wang2020ss, wang2020score}.

Different model-dependent techniques have been proposed to alleviate the poor coverage of objects \cite{lee2019ficklenet, rahimi2020pairwise, xue2019danet, yang2020combinational, zhang2020rethinking}. In \cite{wei2018revisiting}, authors utilized CAMs from multiple convolution layers with different receptive fields to enlarge the discriminative regions. \cite{yang2020combinational} combines CAMs of different classes to identify foreground and background regions in an image. \cite{wu2021background, zhu2021background} propose to suppress background regions to help to generate a foreground map with high confidence, guided by area constraints. For a robust training, \cite{meng2021foreground} incorporates an encoder-decoder layer between the shallow layers, and a generator to mask a part of the image. Low-level feature based activation map (FAM)~\cite{xie2021online} utilizes multiple classifiers to generate a foreground map that is decomposed into multiple part regions. This is used to train a network to produce a final semantic-agnostic map. Self-produced guidance (SPG) \cite{zhang2018self} separates foreground and background regions for guiding shallow learning while expanding to less discriminative areas. Shallow feature-aware pseudo supervised object localization (SPOL) \cite{wei2021shallow} employs a multiplicative fusion strategy for harvesting highly confident regions, and then uses those masks as pseudo-labels for training a segmentation network. Inter-image communication (I$^2$C) \cite{zhang2020inter} increases the robustness of localization maps by considering the correlation of similar images within a class. Structure-preserving activation (SPA) \cite{pan2021unveiling} seeks to preserve the object's structure within a CAM, although this method still produces bloby boundaries. Similarly, Strengthen learning tolerance (SLT) \cite{guo2021strengthen} groups images of similar classes to increase the network's tolerance, and force the localization map to expand towards the object's boundaries. Also, the full resolution CAM (F-CAM) \cite{belharbi2022fcam} has been proposed to expand activation maps to cover full object beyond discriminative regions. Compared to F-CAM, our proposed method is able to define sampling regions for building efficient pseudo-label with the help of classifier. This helps our model to focus only on target object instead of expanding the activation to other objects in the concerned image.

Most of the above methods rely on internal activations acquired by utilizing activation maps at different layers of the networks. Given their intrinsic properties (e.g., local receptive field), CNNs decompose an object into local semantic elements \cite{bau2017network, zeiler2014visualizing}. These intrinsic properties prevent the CNN from forming global relationships between different receptive fields, limiting their ability to localize whole object, and producing coarse bloby maps focused on discriminant regions. One solution to form long-range dependencies is to find global cues by calculating pixel-similarities \cite{wang2018weakly, wang2020self, zhang2020rethinking, zhang2020inter}.  More recently, \cite{ki2020insample} proposed a method based on non-local attention blocks to build long-range relationships and efficiently localize objects. It enhances attention maps by incorporating spatial similarity. 

\noindent\textbf{Vision Transformer (ViT) Methods:} Numerous transformer based methods have recently been proposed for WSOL \cite{bai2022weakly,chen2022lctr,gupta2022vitol,li2022caft,su2022re}. A pioneering work for object localization is the  token semantic coupled attention map (TS-CAM) \cite{gao2021ts}, which is capable of localizing objects at a fine-grained level by fusing attention maps with semantic-aware maps. (The detailed working TS-CAM model is presented in supplementary material.) Recently, \cite{su2022re} proposed a re-attention mechanism for suppressing background regions. Layer-wise relevance propagation \cite{gupta2022vitol} and clustering-based \cite{chen2022lctr} methods have also been proposed to accumulate attention maps produced by ViTs. For instance, \cite{gupta2022vitol} introduces a patch-based attention dropout layer in the attention block, and incorporate an attention roll-out method to improve localization performance. 

Transformer-based methods have had a significant impact on WSOL literature. However, they accumulate all attention maps from each layer, leading to a noisy CAM, or a suppression of different regions within the object \cite{gao2021ts} (see Fig.\ref{fig:dino_last_head}). Most techniques provide a heat map without sharp boundaries. In contrast, our approach can produce high-resolution maps with sharper boundaries, only using attention from the last layer of the transformer block. Methods based on pseudo-labels fix their map for all epochs \cite{wei2021shallow}, generating a map very close to the pseudo-labels. Our method samples only a few pixels per epoch to learn foreground and background maps that continue to evolve based on a bi-nominal distribution. Sampling only a few pixels as pseudo-labels helps the network to explore different parts of the object, covering the whole object with sharper boundaries. Since these methods are typically computationally intensive during training, we harvest maps in a self-supervised way using a pretrained transformer, allowing us to obtain clues about potential objects and train our localization network. Our method leverages long-range dependencies as pseudo-labels are harvested from a transformer, and convolution inductive bias to suppress background noise. Finally, in contrast with WSOL baselines, our method focuses on drone-based WSOL for surveillance of cell tower sites. These images are captured at a long distance, and the object of interest covers a small proportion of the image.

\section{Proposed Method}\label{sec:method}

%

\subsection{Background}\label{sec:background}

\textbf{Vision transformers} (ViTs) are recognized for their success in image classification \cite{dosovitskiy2020image, touvron2021training} and WSOL \cite{gao2021ts}. They consist of $T$ cascaded encoder blocks, each containing multi-headed attention, followed by a multi-layer perceptron (MLP). First, an input image $x$ of size $W \times H$ is divided into $N$ patches of resolution $(W/S)\times (H/S)$, where $S$ denotes the patch size in pixels. These patches are then linearly projected to a fixed embedding size $D$ along with a learnable \cls token 
and passed through $T$ cascaded blocks. Furthermore, the features of a \cls token from $T^{th}$ block are fed to the MLP for class prediction, defined as $p=Softmax_{\tau=1}(MLP(\hat{T}_{vit}(x))$, where 
\begin{equation}
Softmax_{\tau}(s) = \frac{\exp(s / \tau)}{\sum_{k=1}^K \exp(s_k / \tau)}
\end{equation}\label{eq:softmax}
where $s \in \mathbb{R}^K$ is the MLP output, $K$ corresponds to the number of classes, and $\hat{T}_{vit}$ represents the transformer's forward pass, generating features $f$ that are passed through $T$ cascaded blocks. The MLP  predicts the probabilities of concerned classes with cross-entropy classification loss $- \Sigma_y y log(p_y)$, and $y$ is the ground truth label. 



\textbf{Self-\textit{Di}stillation with \textit{no} labels (DINO)} \label{sec:dino} involves harvesting regions of interest without class-level labels. To achieve this goal, the authors trained two networks, called student and teacher, to match their probability distribution. Parameters of the teacher are an exponential moving average of the student network. The probability $p$ of the representation for these networks is calculated by normalizing the output of the concerned network using the softmax function with temperature defined in Eq. \ref{eq:softmax} ($\tau$ is a hyper-parameter to be optimized). Furthermore, the input $x$ is augmented in different ways, $x_1$ and $x_2$, for the student and teacher networks, respectively. They represent two set of distorted views of image $x$, calculated using the strategy defined in \cite{caron2020unsupervised}. The first set contains several local views, while the second one contains two global views. Here, local views contain less than $50\%$ of the original image, and global views cover more than $50\%$ of the area. Finally, the loss is calculated as cross-entropy between the output probability of the student $p^s$ and the teacher $p^t$:
\begin{equation}
        \min_{\theta_s} -p^s(x_1) \log( p^t(x_2) ). 
  \label{eq:kd}
\end{equation}

\begin{figure*}[!t]
\begin{center}
 \includegraphics[width=0.94\linewidth,trim=4 4 4 4, clip]{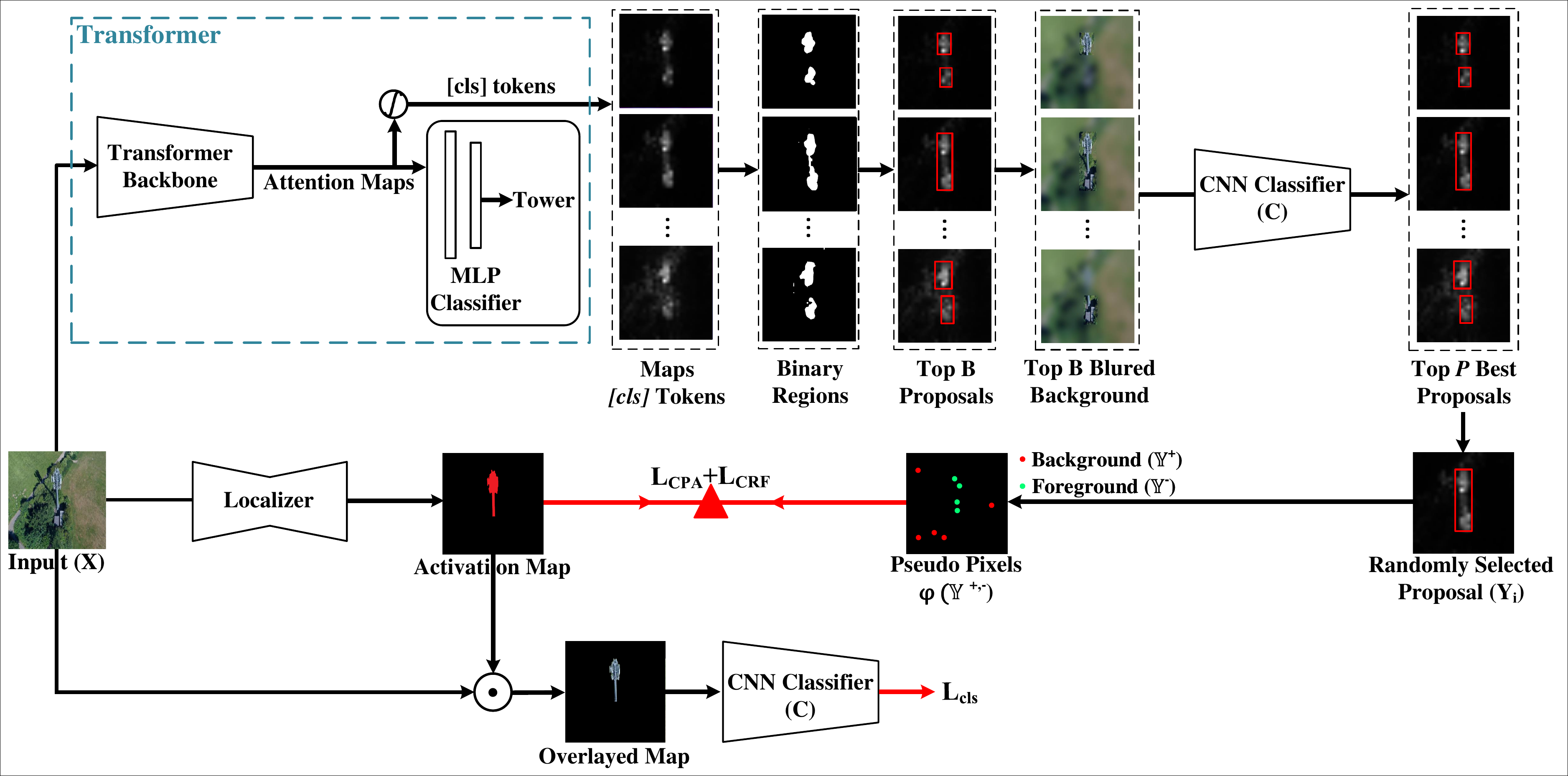}
\end{center}
\caption{Architecture for training a deep WSOL network using the DiPS method.  
A pertained transformer is employed along with a classification layer with frozen weights. After obtaining attention maps, a threshold is applied to produce binary proposals and generate bounding boxes for each proposal. Based on those proposals, areas outside the box are blurred, and the top \textit{P} best bounding boxes are selected based on the classifier's score that are greater than a minimum score. Ultimately, a bounding box is selected among the \textit{P} best boxes for sampling from background and foreground regions. Then, the generated pseudo-pixels are considered as pseudo-labels to training of the localization network. After training, an input images are processes localization network without the need of a transformer's backbone. 
}\label{fig:our_method}
\end{figure*}

\subsection{Overview of DiPS}

As we discussed, SSTs \cite{caron2021emerging} are able to localize all objects in an image. They decompose the objects in an image into different maps because they learn to represent objects in a self-supervised fashion.  Each map highlights different objects or their parts without associating categorical information with them. If we can successfully identify maps and corresponding regions that contain the concerned object, then we can use them as pseudo-labels to train a localization network. Therefore, we propose a new DiPS method that can harvest efficient pseudo-labels on the target object for accurate WSOL. DIPS utilizes \cls tokens of a pretrained ViT to obtain effective pseudo-labels and a pretrained CNN classifier to select ROIs proposals to train our localization network (see Fig.\ref{fig:our_method}). It allows us to directly minimize the loss over the generated map for accurate localization maps with a similar confidence for all object parts. 

Given a set of tokens from a pretrained SST model \cite{caron2021emerging}, the associated maps are first binarized, allowing to extract bounding boxes over potential ROIs. They are considered proposal candidates that potentially cover object regions with varying certainty. To measure certainty, the response of a pretrained CNN classifier is employed to indicate the most relevant ROI. To avoid overfitting to a single proposal, we consider top-$P$ confident ROI proposals, and randomly select one of them for sampling. This final proposal is used to constrain sampling of pixel-wise pseudo-labels where foreground pixels are sampled inside the bounding box of the proposals. Sampling is guided by the activation magnitude, where strong activations are favored to be foreground. Background pixels are sampled outside the bounding box by favoring low activations. Pseudo-labels are sampled randomly at each Stochastic Gradient Descent (SGD) iteration, which has been shown to be more effective and robust against overfitting than static and noisy pseudo-labels~\cite{negevsbelharbi2022,belharbi2022fcam}. Overall, this sampling allows for the emergence of accurate localization maps, from a few pixels toward segments and the entire object. This can be seen as a \emph{fill-in-the-gap} game where only random pixels are selected as foreground/background, and the localization network must generalize to similar regions. Then, over iterations, this random sampling is expected to stimulate convolution filters to have a similar response to visually similar nearby pixels.

The collected pixel-level pseudo-labels are used to train a U-Net style localization network~\cite{ronneberger2015u}, as illustrated in Fig.\ref{fig:our_method}. It takes the image as input and yields foreground and background CAMs for localization. To train this model, we consider a composite loss that leverages local and global constraints. Local constraints aim to guide learning at pixel level. To this end, standard partial cross-entropy is used to exploit sampled pixel-wise pseudo-labels. To produce a consistent CAM that is well aligned with object boundaries, we also include a Conditional Random Field (CRF) loss~\cite{tang2018regularized}. To further ensure that the localized object is well aligned with the image-class label, we add a global constraint over the foreground CAM. In particular, we feed the product of the image by the foreground CAM to the pretrained classifier. Then, the classifier response is maximized with respect to the true image-class. During training, only the weights of the localization network are optimized. The SST and classifier models are frozen. At the end of the training, only the localization network and classifier models are retained for classification and localization tasks. 
The rest of this section provides more details on DiPS components.


\subsection{Training Architecture}
 In this paper, our goal is to localize the object of interest through an encoder-decoder localization network that is trained using pseudo-labels. These pseudo-labels are obtained from \cls token of the different heads from the last layer of the transformer. 
 We represent the training set by $\mathbb{T} = \{\mathrm{x}_i, \mathrm{y}_i\}$ and 
$\mathrm{x}\in R^{W\times H}$ represents an image with a size of $W\times H$ and its corresponding label of $K$ classes is represented by $y\in\{1,\dots,K\}$. DiPS employs a mechanism to sample background and foreground regions from activation maps (\cls token) to build effective pseudo-labels, as shown in Figure \ref{fig:our_method}. To achieve these goals, we propose a model that consists of three modules (i) transformer $\slantbox{$\mathrm{T}$}_{SST}$ that was trained in self-supervised fashion (as explained above) for producing an attention map (ii) localization network $\slantbox{$\mathrm{F}$}_\theta$ for producing soft-max activation maps represented by $M=\slantbox{$\mathrm{F}$}_\theta(x)\in[0,1]^{n\times m\times 2}$; here first channel represent the foreground $M_1$ and other represent the background map $M_2$ trained using best proposal extracted from \cls tokens (iii) a classification layer $\slantbox{$\mathrm{\hat{C}}$}$ that is trained on top of transformer's features that is used to calculate additional matrices that involves class accuracy (see supplementary material). 
    
    
    \noindent \textbf{Selection of Background and Foreground Pixels:} In order to train the localization map predictor a pixel-level supervisory signal is employed. At each epoch, new pseudo-labels are generated that consist of a few foreground and background pixels (pseudo-pixels) to increase network certainty of different regions during training.  To select pseudo-pixels, we extract attention maps from the last layer of the transformer, corresponding to \cls tokens. 
    From the maps, we select only the first four maps and an average map of all \cls tokens. 
    After this, we apply the Otsu \cite{otsu1979threshold} thresholding method for converting the attention map into a binary map to obtain all connected regions that are greater than a minimum size\footnote{The minimum size is an hyperparameter value.} to extract all $B$ proposal (bounding) boxes enclosing those regions. 
    Then, for an input image, we produce $B$ images corresponding to each bounding box by applying the Gaussian-blur filter to hide the background regions outside of the respective bounding box. Each image is then passed to a classifier $\slantbox{$\mathrm{C}$}$, and the top \textit{P} best proposals along with their underlying attention maps are selected based on the classifier's confidence corresponding to target class $\mathrm{y}_i$. Then, one out of \textit{P} proposals along with its respective bounding box is selected for pseudo labeling. The pixels inside and outside of the bounding box are considered as foreground \slantbox{$\mathbb{Y}_i^+$} and background regions \slantbox{$\mathbb{Y}_i^-$} respectively defined as:
    \begin{equation}\label{loss:fg_bg_pixels}
        \slantbox{$\mathbb{Y}_i^+$}=\varphi^+(Y_i,n^+) \quad\quad\quad \slantbox{$\mathbb{Y}_i^-$}=\varphi^-(Y_{i},n^-)
    \end{equation}

\begin{figure*}[!t]
\centering
		\includegraphics[width=1\linewidth,trim=0 0 0 0,clip]{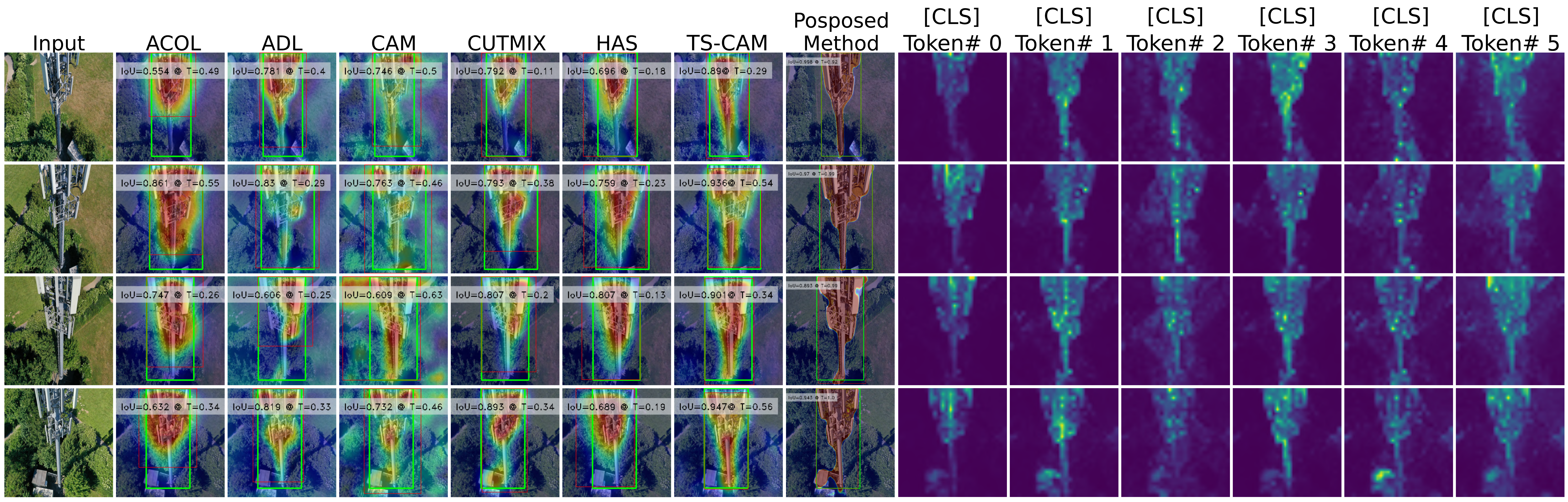}
		\caption{Visualization of test samples from TelDrone. Results of baseline method including TS-CAM shows that these method can only cover some parts of the object and invisible parts are also included in the bounding box due to extensive search of thresholds. In contrast, our method learns to predict map that can cover the whole object. Moreover, maps in red box are used to sample pseudo-labels and our methods learns to localize the object precisely and able remove noise from them. }
		\label{fig:results_teldrone}
\end{figure*}

\begin{figure*}[!b]
\begin{center}
\includegraphics[width=1\linewidth,trim=0 0 0 0,clip]{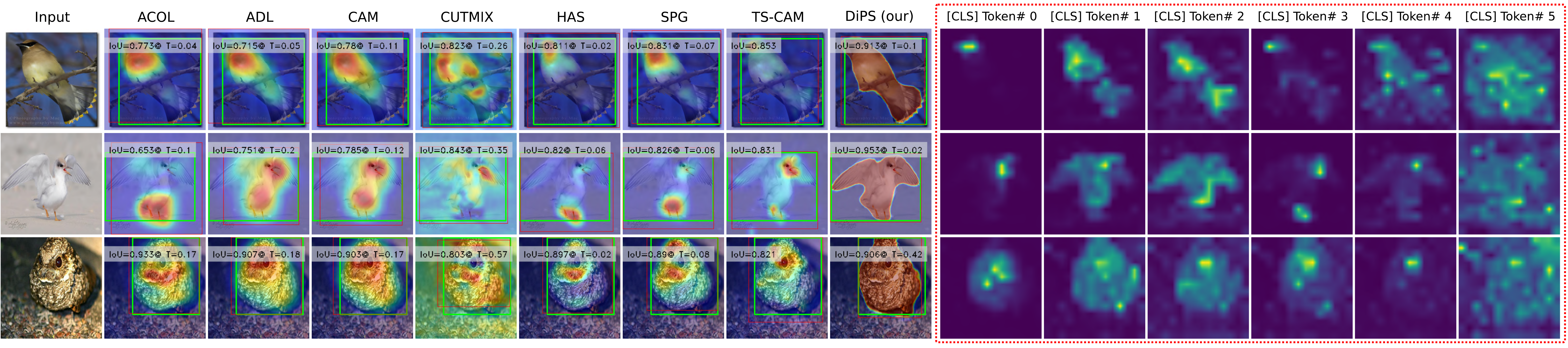}
\end{center}\caption{Visualization of test samples from CUB dataset.}\label{fig:results_cub}
\end{figure*}

\noindent where $Y_i$ set of all pixel in the generated map, \slantbox{$\mathbb{Y}_i^+$} represents the top $n\%$ pixels within the bounding box selected through multinomial sampling and \slantbox{$\mathbb{Y}_i^-$} represents top $-n\%$ pixels anywhere that are derived using uniform sampling from the activations of whole image sorted in reverse order. These pixels are derived from unreliable attention maps and may contain incorrect labels. To deal with this uncertainty, we select only a few (see experimental details for exact values) pixels from the foreground and background while rejecting others \cite{singh2017hide, srivastava2014dropout}. Therefore, the final set of foreground and background pixels for calculating loss are:
    \begin{equation}\label{loss:fg_bg_union}
        \varphi(\slantbox{$\mathbb{Y}$}_i^{+,-}) = 
        \varphi(\slantbox{$\mathbb{Y}_i^+$})\cup\varphi(\slantbox{$\mathbb{Y}_i^-$})
    \end{equation}
    here $\varphi(\slantbox{$\mathbb{Y}_i^+$})$ and $\varphi(\slantbox{$\mathbb{Y}^-$})$ represent set of pixels (hyper-parameter) sampled using multinomial distribution. At the end, we then generated pseudo label $Y\in\{0,1\}^2$ by assigning 1's to pixel in \slantbox{$\mathbb{Y}_i^+$} and 0's to pixels in set \slantbox{$\mathbb{Y}_i^-$}. 
    
    \noindent \textbf{Overall Training Loss:} { Our training loss consists of three terms; (i) classifier's loss to ensure the integrity of generated map. The generated map $M_1$ is overlaid on the image $x$ and passed it to the classifier to compute a cross-entropy loss; $\mathcal{L}_{CLS}=-\sum_{k=1}^{K}y_klog(Softmax(\slantbox{$\mathrm{C}$}{(x\odot M_1)})_k)$.} (ii) Our constrained pixel alignment loss\footnote{Loss using few pixels was also employed by \cite{belharbi2022fcam}.} for learning foreground/background regions. It aligns the output map $M$ with the selected pixels in $\varphi(\slantbox{$\mathbb{Y}$}_i^{+,-})$ through partial cross-entropy denoted by $\mathcal{L}_{CPA}(\varphi(\slantbox{$\mathbb{Y}$}_i^{\;r}), M^{\;r})$; here $r$ represent the selected pixels. (iii) Conditional Random Field (CRF) \cite{tang2018regularized} to align the localization map with the object boundaries. The detailed description of the CRF loss $\mathcal{L}_{CRF}$ is presented in supplementary material. Furthermore, the overall can be formulated as,
    \begin{equation}\label{loss:confidence}
        \mathcal{L}_{Total} = \min_{\theta} \lambda_{CLS}\mathcal{L}_{CLS} + \lambda_{CPA} \mathcal{L}_{CPA}  + \lambda_{CRF} \mathcal{L}_{CRF}
    \end{equation}
    where $\lambda_{CLS}$ and $\lambda_{CPA}$ are hyperparameters between interval [0, 1], and $\lambda_{CRF}$ is set to $2e^{-9}$ as defined in \cite{tang2018regularized}.
\section{Results and Discussion}\label{sec:exp}

\subsection{Experimental Methodology} \label{sec:expsetting}


\noindent \textbf{Datasets:} 
For validation, we employed two datasets: CUB-200-2011 and an internal dataset named TelDrone.  
\textbf{CUB200-2011} is one of the most popular datasets used for visual classification and object localization tasks. This dataset consists of 11,788 images divided into 200 categories; 5,794 for testing and 5,994 for training \cite{WelinderEtal2010}. For validation and hyperparameter search, we employed an independent validation set collected by \cite{choe2020evaluating} containing 1,000 images. \textbf{TelDrone} contains 915 4K images collected using a drone that orbits around the tower site. Furthermore, images are divided into two classes (either containing an inspection site or not). From this dataset, we considered 797 images for training, 13 images for validation and 105 images as a test set.

\noindent \textbf{Evaluation measures:}
We analyze our results based on the evaluation measure suggested in \cite{choe2020evaluating} -- \newmaxboxacc refers to the proportion of predicted bounding boxes that have an IoU greater than a particular threshold concerning the generated map. It is averaged over three different IoU thresholds $\delta \in \{30\%, 50\%, 70\%\}$. We also reported additional localization matrices in supplementary material as presented in \cite{choe2020evaluating}. 




\noindent \textbf{Implementation details:}
We follow the protocol in \cite{choe2020evaluating} for all experiments. 
A batch size of 32 is used for all datasets. 
For training on CUB, images are resized to $256\times256$, and then randomly cropped to $224\times224$ and randomly flipped horizontally as in \cite{choe2020evaluating}. Moreover, for the TelDrone dataset, we first resize images to $512\times512$ followed by random cropping of size $448\times448$. 
We trained our model for 50 epochs while the learning rate is decayed by a factor of 0.1 after every $15^{th}$ epoch for both dataset.

\noindent \textbf{Baseline Models:} To evaluate the performance of our model, we compared it with several state-of-the-art methods presented in Tables \ref{tab:openimages}. We report results from \cite{choe2020evaluating} for CAM \cite{zhou2016learning}, HaS \cite{singh2017hide}, ACoL \cite{zhang2018adversarial}, SPG \cite{zhang2018self}, ADL \cite{choe2019attention}, and CutMix \cite{yun2019cutmix}. 
For all other models, we report the same results as published in their respective articles. For qualitative evaluation, we reproduced the results of CAM \cite{zhou2016learning}, HaS \cite{singh2017hide}, ACoL \cite{zhang2018adversarial}, SPG \cite{zhang2018self}, ADL \cite{choe2019attention}, CutMix \cite{yun2019cutmix} and TS-CAM \cite{gao2021ts} according to the protocols defined in \cite{choe2020evaluating}. 

\subsection{Comparison with State-of-Art Methods}

Table \ref{tab:openimages} shows that DiPS achieves state-of-the-art performance on CUB and TelDrone datasets for \newmaxboxacc metric compared to the other methods in the literature. The visual results of our method along with the corresponding baselines on TelDrone and CUB datasets are presented in Fig.\ref{fig:results_teldrone} and Fig.\ref{fig:results_cub} respectively. 
The baseline method tends to focus on discriminative regions and expands to less discriminative regions because of the bloby nature of these maps. However, the selection of the optimal threshold allows the less discriminative area to be included in the localization map because of their texture or color similarity. Furthermore, the resultant map becomes unreliable and unable to identify the whole object as it encompasses regions with no concealed activation over an object.
For CUB dataset, we compare the results with the baseline methods (Fig.\ref{fig:results_cub}) including TS-CAM and show that our method is capable of covering the whole object with the same intensities instead of highlighting different parts of the object. Additionally, the \cls tokens of SST are also presented, and DiPS is able to successfully harvest useful information from them by constructing a reliable pseudo-label. Similarly, on the TelDrone dataset, DiPS achieved state-of-the-art visual results 
by localizing all parts of the object with the same confidence instead of hot-spotting different regions (Fig.\ref{fig:results_teldrone}). 
More specifically, current state-of-the-art methods (e.g. TS-CAM) focus on some regions of the concerned object by hot-spotting different parts. Due to this they are able to properly draw a bounding box around the object due to extensive threshold search that can include low scoring areas to the localization map.
In contrast, our method can produce a map that covers the whole object with sharper boundaries, which eliminates the need for a precise threshold value. 
Additionally, we compare the results of our model with SST's activation and show that our method learns to localize well even with noisy pseudo-labels. 

\begin{table}[!h]
\centering
\resizebox{1\linewidth}{!}{%
\centering
\begin{tabular}{lcccc}

\textbf{Methods} &  & \textbf{TelDrone} && \textbf{CUB}\\
 &  & (\textbf{\newmaxboxacc}) && \textbf{(\newmaxboxacc)}\\
\cline{3-3} \cline{1-1} \cline{3-3} \cline{5-5}  
CAM~\cite{zhou2016learning} {\small \emph{(cvpr,2016)}} & & 55.9 && 63.7 \\
HaS~\cite{singh2017hide} {\small \emph{(iccv,2017)}} & & 60.3 && 64.7 \\
ACoL~\cite{zhang2018adversarial} {\small \emph{(cvpr,2018)}} & & 59.1 && 66.5 \\
SPG~\cite{zhang2018self} {\small \emph{(eccv,2018)}} & & 67.3 && 60.4 \\
ADL~\cite{choe2019attention} {\small \emph{(cvpr,2019)}} & & 66.0 && 66.3\\
CutMix~\cite{yun2019cutmix} {\small \emph{(eccv,2019)}} & & 57.2 && 62.8\\

ICL~\cite{ki2020insample} {\small \emph{(accv,2020)}} & & -- && 63.1 \\


TS-CAM~\cite{gao2021ts} {\small \emph{(iccv,2021)}} & & 72.2 && 76.7 \\
CAM-IVR~\cite{kim2021normalization} {\small \emph{(iccv,2021)}} & & -- && 66.9 \\
PDM~\cite{meng2022diverse} {\small \emph{(tip,2022)}} && -- & & 72.4 \\
C$^2$AM~\cite{xie2022c2am} {\small \emph{(cvpr,2022)}} & & -- && 83.8 \\

ViTOL-GAR~\cite{gupta2022vitol} {\small \emph{(cvpr,2022)}} & & -- && 72.4 \\
ViTOL-LRP~\cite{gupta2022vitol} {\small \emph{(cvpr,2022)}} & & -- && 73.1 \\
TRT~\cite{su2022re} {\small \emph{(corr,2022)}} & & -- && 82.0 \\
BGC~\cite{kim2022bridging} {\small \emph{(cvpr,2022)}} & & -- && 80.1 \\
SCM~\cite{bai2022weakly} {\small \emph{(eccv,2022)}} & & -- && 89.9  \\
CREAM~\cite{xu2022cream} {\small \emph{(cvpr'22)}} & & -- && 73.5 \\
F-CAM+CAM~\cite{belharbi2022fcam} {\small \emph{(wacv'22)}} & & -- && 79.4 \\
F-CAM+LayerCAM~\cite{belharbi2022fcam} {\small \emph{(wacv'22)}} & & -- && 80.1 \\
\cline{1-1}\cline{3-3}\cline{5-5}
\textbf{DiPS (ours)} & & 84.9 && 90.9 \\
\cline{1-1}\cline{3-3}\cline{5-5} \\
\end{tabular}
}
\vspace{0.6em}
\caption{\newmaxboxacc performance on the TelDrone and CUB.}
\label{tab:openimages}
\end{table}

\begin{figure}[!h]
\begin{center}
 \includegraphics[width=0.8\linewidth,trim=3 3 3 3, clip]{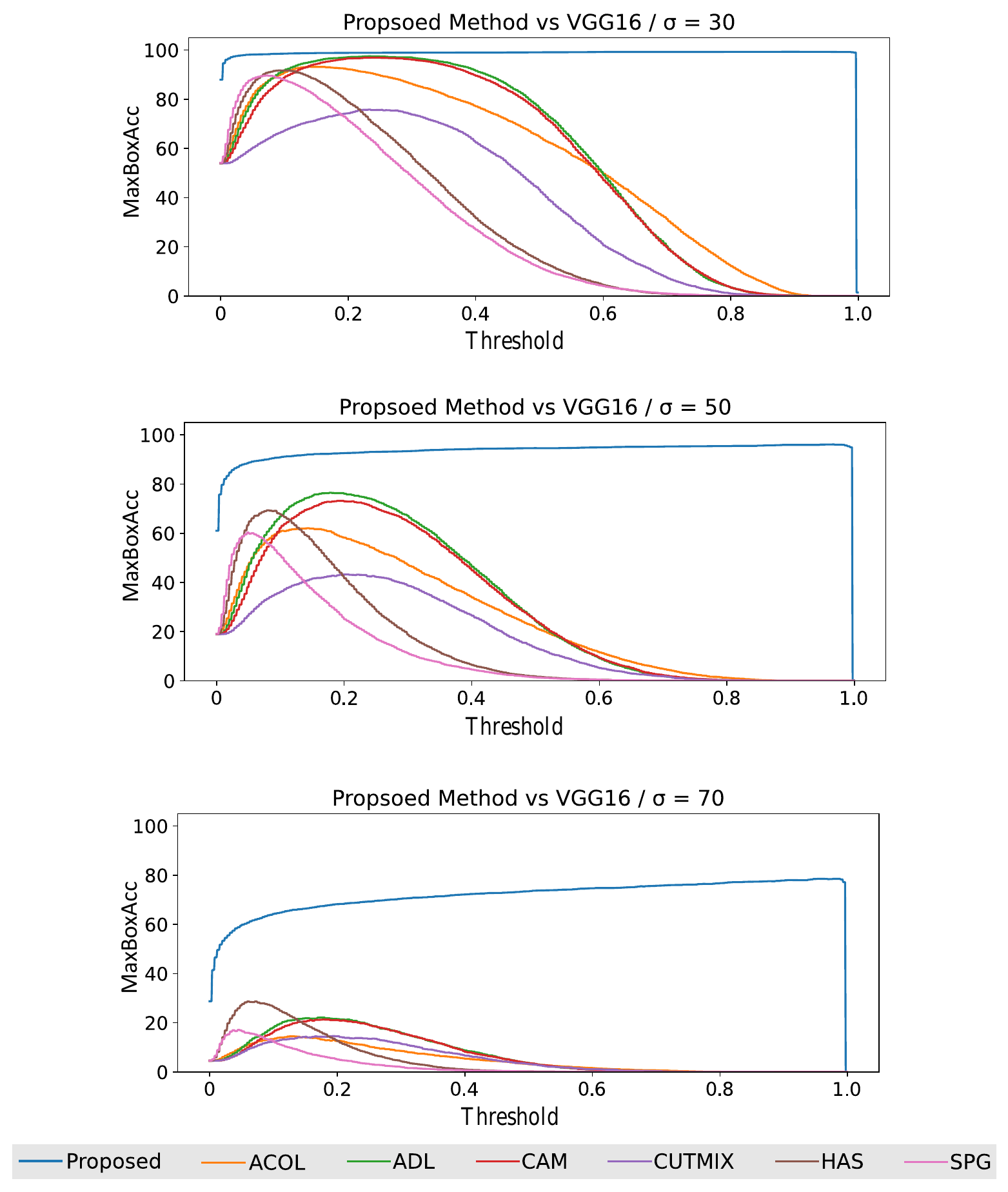}
\end{center}
\caption{\maxboxacc performance at different IoU threshold values for output map using the CUB dataset.}\label{fig:acc_over_ths}
\end{figure}

\noindent \textbf{Analysis of distribution shift.}
We present the effect of varying the threshold values on the localization performance along with the distribution spread of foreground and background regions. The change in \maxboxacc at different threshold for CUB test set is presented in Figure \ref{fig:acc_over_ths}. We compare our results with CAM \cite{zhou2016learning}, HaS \cite{singh2017hide}, ACoL \cite{zhang2018adversarial}, SPG \cite{zhang2018self}, ADL \cite{choe2019attention}, and CutMix \cite{yun2019cutmix}. For all of these related methods, we can see that the \maxboxacc quickly drops to zero as the threshold increases. Due to this, it is very difficult to find the optimal threshold for each image which makes these methods unfeasible to deploy in practical scenarios. In contrast, the output of our method is less susceptible to the threshold.

\section{Conclusion}\label{sec:conc}
In this paper, we proposed a DL method for WSOL, capable of producing fine-grained object localization maps of cell tower sites in aerial images for assets surveillance. The proposed method is capable of learning from pseudo-labels instead of only class labels. These pseudo-labels are harvested by sampling foreground and background areas from the \cls token of an SST. Our method is able to generate better maps compared to the \cls token used to build pseudo-labels. The proposed method is capable of generating reliable maps, thus providing better coverage of the whole object. 
To validate the performance of our model, we compared its results with the existing baseline method, and our method achieved competitive performance both 
qualitatively and quantitatively. Compared to the baseline, our method can identify all parts of the object instead of decomposing and highlighting different parts of the objects. Furthermore, the proposed method eliminates the need for an extensive threshold search to produce an optimal bounding box covering the concerned object. 

\noindent \textbf{Note:} Additional details are provided in the supplementary materials, including additional visual results, an evaluation matrix for error dissection, an extended error analysis, ablation studies, and an overview of the CRF loss.


\noindent \textbf{Acknowledgments:} This research was supported by MITACS and the Natural Sciences and Engineering Research Council of Canada. We also acknowledge Compute Canada for the computing resources.

\appendix
\newcommand{\hbAppendixPrefix}{A}

\section{Supplementary Material}

\renewcommand{\thefigure}{\hbAppendixPrefix\arabic{figure}}
\setcounter{figure}{0}
\renewcommand{\thetable}{\hbAppendixPrefix\arabic{table}} 
\setcounter{table}{0}
\renewcommand{\theequation}{\hbAppendixPrefix\arabic{equation}} 
\setcounter{equation}{0}

\noindent This section contains the following material:
\begin{enumerate}[a]
    \itemsep0em 
    \item [A.1] an evaluation measures and error dissection;
    \item [A.2] an overview of CRF loss;
    \item [A.3] additional error analyses;
    \item [A.4] ablation studies;
    \item [A.5] additional visual results;
    \item [A.6] details on the TS-CAM method.
\end{enumerate}

\subsection{Performance Measures and Error Dissection}

\subsubsection{Evaluation Measure for Error Dissection}\label{app:errordiss}
In this section, we present the evaluation measures that are used in \cite{gao2021ts} for error dissection over wrong predictions. These measures are useful for deciding threshold values for producing bounding boxes from localization maps. Specifically, localization part error (\lpe) and localization more error (\lme) help in deciding whether to increase or decrease the threshold value for optimal results. More details on error measures are given below: 

\noindent \textbf{Localization Part Error (\lpe): } This measure identifies that an object partially detected by the localization map with a large margin has an intersection over the predicted bounding box (\iop) $>$ 0.5.

\noindent \textbf{Localization More Error (\lme): }
It indicates that the predicted bounding box is larger than the actual box and covers other objects or background. This can be identified if intersection over annotated-bounding-box (\ioa) $>$ 0.7.


\subsubsection{Additional Performance Measures}\label{app:accdiss}
\noindent \textbf{\ltopone Localization Accuracy (\ltopone Loc): } A prediction is considered true if the predicted class is the same as the ground truth and the intersection over Union (\iou) is greater than 0.5.

\noindent \textbf{\ltopfive Localization Accuracy (\ltopfive Loc): } A prediction is considered true if the \iou is greater than 0.5 and the actual class matches at least one of the top 5 predicted classes.

\subsection{Overview of CRF loss}

Conditional random fields (CRF) loss, aligns the predicted localization map $\mathrm{M}$ with the boundaries of a concerned object presented in input $\mathrm{x}$. CRF loss \cite{tang2018regularized} between $\mathrm{x}$ and $\mathrm{M}$ can be defined as follows: 
\begin{equation}\label{loss:crf_loss}
        \mathcal{L}_{CRF} (\mathrm{A}, \mathrm{M}) = \sum_{i=0}^{i=1} {M_{i}}^T A (1-M_{i})
\end{equation}
where $\mathrm{A}$ represents an affinity measure that contains mutual similarities between pixels, including proximity and color information. For capturing affinities of pixels, we use a Gaussian kernel \cite{koltun2011efficient} and employ permutohedral lattice \cite{adams2010fast} to reduce the computation overhead.

\begin{table}[!h]
\centering
\resizebox{0.5\linewidth}{!}{
\begin{tabular}{lccccccc}
     & \lpe$\downarrow$ & \lme$\downarrow$ 
     \\
     \cline{2-3}
    VGG16  &21.91 &10.53 
    \\
    InceptionV3 &23.09 &5.52 
    \\
     TS-CAM \cite{gao2021ts}  & 6.30 &2.85  
     \\
     \hline
     DiPS (our) & 0.05  & 0.07 
     \\ 
	\bottomrule
\end{tabular}
}
\vspace{0.5em}
\caption{Extended error analysis on the CUB-200-2011 dataset}
\label{tab:err_analysis_cub}.
\end{table}
\subsection{Extended Error Analysis}
Further error analysis (according to the error measures defined in Section \ref{app:errordiss}) on the CUB datasets is presented in Table \ref{tab:err_analysis_cub}. Our method localized the correct region of the concerned object instead of overestimating or underestimating the region. 
It also shows that the maps generated by DiPS are very robust and have much fewer errors compared to the baseline methods. The statistics of the baseline methods are from \cite{gao2021ts}.

    



\subsection{Additional Ablation Studies}
The performance of DiPS with various loss function combinations is shown in Table \ref{tab:ablation_study_cub_teldrone}. It shows that all of the auxiliary losses contribute significantly towards the final performance. Also, training through arbitrary selection of pixels (pseudo-labels) rather than the classifier loss or fixed pseudo-labels allows DL models to explore different regions of an object and can provide accurate localization. Adding CRF and classification terms at the same time significantly improves the performance of our model. The \maxboxacc of our model on TelDrone is presented in table \ref{tab:cub_teldrone}. Furthermore, the \maxboxacc, \topone and \topfive localization accuracy for CUB dataset is presented in Table \ref{tab:cub_supp}. We achieved state-of-the-art performance on the TelDrone and CUB dataset. 

\begin{table}[!h]
\centering
\resizebox{0.8\linewidth}{!}{
\begin{tabular}{lcccccc}
     && CUB&& TelDrone\\
     && (\maxboxacc) && (\maxboxacc)\\
    \cline{1-1}\cline{3-3}\cline{5-5}
    $\mathcal{L}_{CPA} + \mathcal{L}_{CRF}$ && 95.4 && 93.3\\
    $\mathcal{L}_{CPA} + \mathcal{L}_{CLS}$ && 94.6  && 91.7\\
    $\mathcal{L}_{CPA} + \mathcal{L}_{CRF} + \mathcal{L}_{CLS}$ && 97.0 && 96.2 \\
	\cline{1-1}\cline{3-3}\cline{5-5}
\end{tabular}
}
\vspace{0.5em}
\caption{Localization performance of our DiPS method with different losses.}
\label{tab:ablation_study_cub_teldrone}
\end{table}

\begin{table}[!h]
\centering
\resizebox{0.7\linewidth}{!}{%
\centering
\small
\begin{tabular}{lcccccc}

\cline{3-5}
& & & \maxboxacc\\
\cline{3-4}
\cline{1-1}\cline{3-5}
CAM~\cite{zhou2016learning} {\small \emph{(cvpr,2016)}} & &  & 50.9  \\
HaS~\cite{singh2017hide} {\small \emph{(iccv,2017)}} & &  & 60.4  \\
ACoL~\cite{zhang2018adversarial} {\small \emph{(cvpr,2018)}} & &  & 62.3 \\
SPG~\cite{zhang2018self} {\small \emph{(eccv,2018)}} & &  & 67.9  \\
ADL~\cite{choe2019attention} {\small \emph{(cvpr,2019)}} & &  & 73.5  \\
CutMix~\cite{yun2019cutmix} {\small \emph{(eccv,2019)}} & &  & 54.7  \\


\cline{1-1}\cline{3-5}
\textbf{DiPS (ours)} & &  & 962 &  &  \\
\cline{1-1}\cline{3-5}
\end{tabular}
}
\vspace{0.6em}
\caption{\maxboxacc performance on the TelDrone dataset.}
\label{tab:cub_teldrone}

\end{table}

\begin{table}[!h]
\centering
\resizebox{0.7\linewidth}{!}{%
\centering
\small
\begin{tabular}{lcccccc}

& & \multicolumn{5}{c}{\textbf{CUB}}\\
\cline{3-6}
& & & \maxboxacc & \topone Loc Acc & \topfive Loc Acc
\\
\cline{3-6}
\cline{1-1}\cline{3-5}\cline{7-7}
CAM~\cite{zhou2016learning} {\small \emph{(cvpr,2016)}} & &  & 73.2 & 56.1 & --  \\
HaS~\cite{singh2017hide} {\small \emph{(iccv,2017)}} & &  & 78.1 & 60.7 & --  \\
ACoL~\cite{zhang2018adversarial} {\small \emph{(cvpr,2018)}} & &  & 72.7 & 57.8 & -- \\
SPG~\cite{zhang2018self} {\small \emph{(eccv,2018)}} & &  & 63.7 & 51.5 & --  \\
ADL~\cite{choe2019attention} {\small \emph{(cvpr,2019)}} & &  & 75.7 & 41.1 & --  \\
CutMix~\cite{yun2019cutmix} {\small \emph{(eccv,2019)}} & &  & 71.9 & 54.5 & --  \\

ICL~\cite{ki2020insample} {\small \emph{(accv,2020)}} & &  & 57.5 & -- & -- \\
TS-CAM~\cite{gao2021ts} {\small \emph{(iccv,2021)}} & &  & 87.7 & 71.3 & 83.8 \\



BR-CAM~\cite{zhu2022bagging} {\small \emph{(eccv,2022)}} & && -- & -- & --\\
CREAM~\cite{xu2022cream} {\small \emph{(cvpr,2022)}} & &  & 90.9 & 71.7 & 86.3 \\
BGC~\cite{kim2022bridging} {\small \emph{(cvpr,2022)}} & &  & 93.1 & 70.8 & 88.0 \\
F-CAM~\cite{belharbi2022fcam} {\small \emph{(wacv,2022)}} & &  & 92.4 & 59.3 & 82.7 \\

\cline{1-1}\cline{3-6}
\textbf{DiPS (ours)} & &  & 97.0 & 78.8 & 91.3 \\
\textbf{DiPS (ours) (w/ TransFG classifier \cite{he2022transfg})} & &  & 97.0 & 88.2 & 95.6 \\
\cline{1-1}\cline{3-6}
\end{tabular}
}
\vspace{0.6em}
\caption{\maxboxacc, \topone and \topfive performance on the CUB dataset.}
\label{tab:cub_supp}

\end{table}

\subsection{Visual Results}
\begin{figure*}[!h]
\begin{center}
 \includegraphics[width=1\linewidth,trim=3 3 3 3, clip]{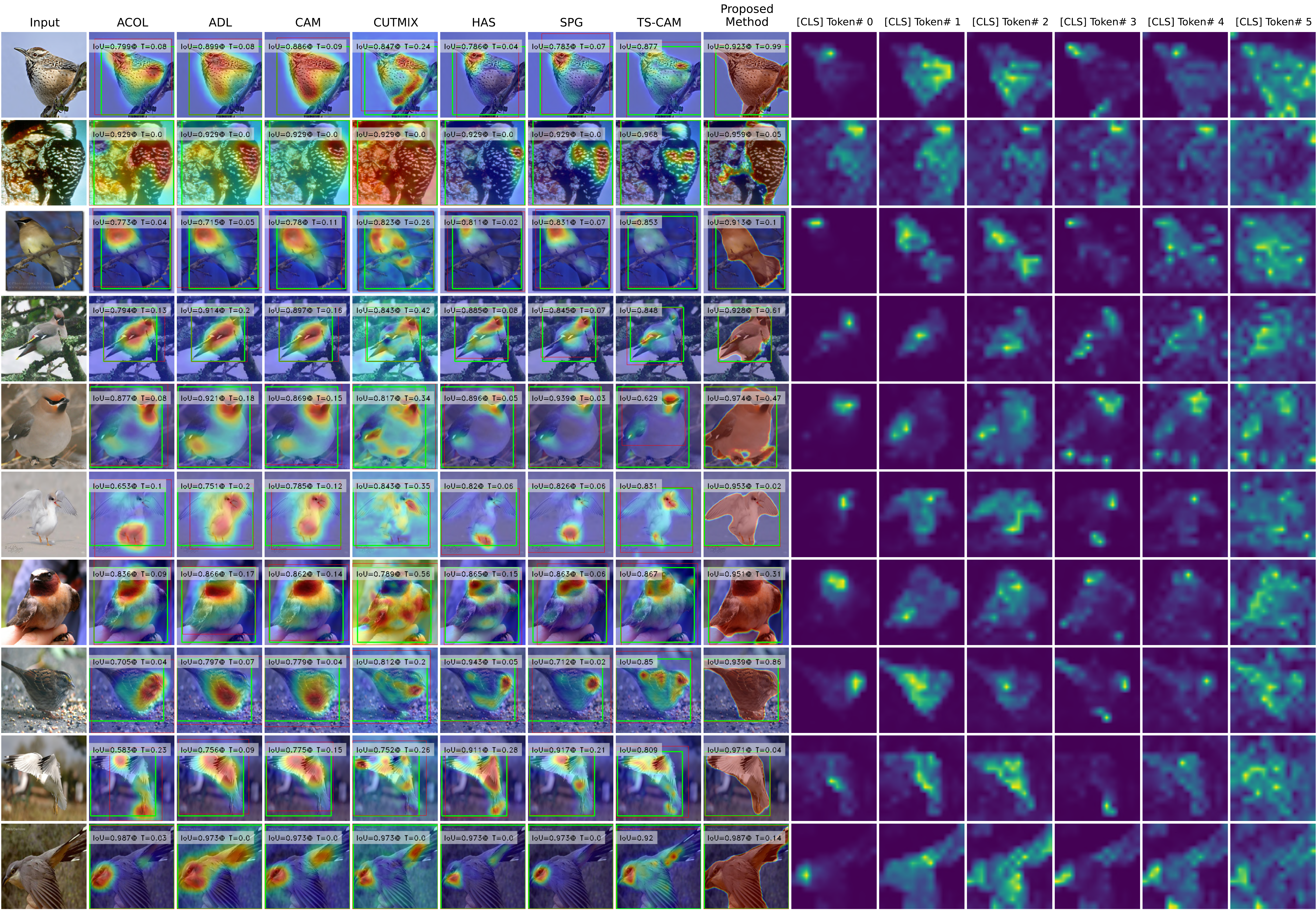}
\end{center}
\caption{Examples of visual results on the CUB-200-2011 dataset.}\label{app:fig_cub_prb_ful}
\end{figure*}

Visual representation of our method compared to baseline methods on CUB is illustrated in Fig.\ref{app:fig_cub_prb_ful}. Our method generates a very smooth map instead of hot-spotting different parts of the concerned object. Ultimately, the map generated by our method does not require an extensive threshold search to find an optimal bounding box. Compared to the \cls tokens of SST (used to harvest pseudo-labels), our method is able to learn an effective localization map from noisy pseudo-labels. 

\subsection{Details on Baseline Method: TS-CAM} \label{sec:relatedmethod}

\begin{figure}[!b]
\begin{center}
 \includegraphics[width=0.73\linewidth,trim=4 4 4 4, clip]{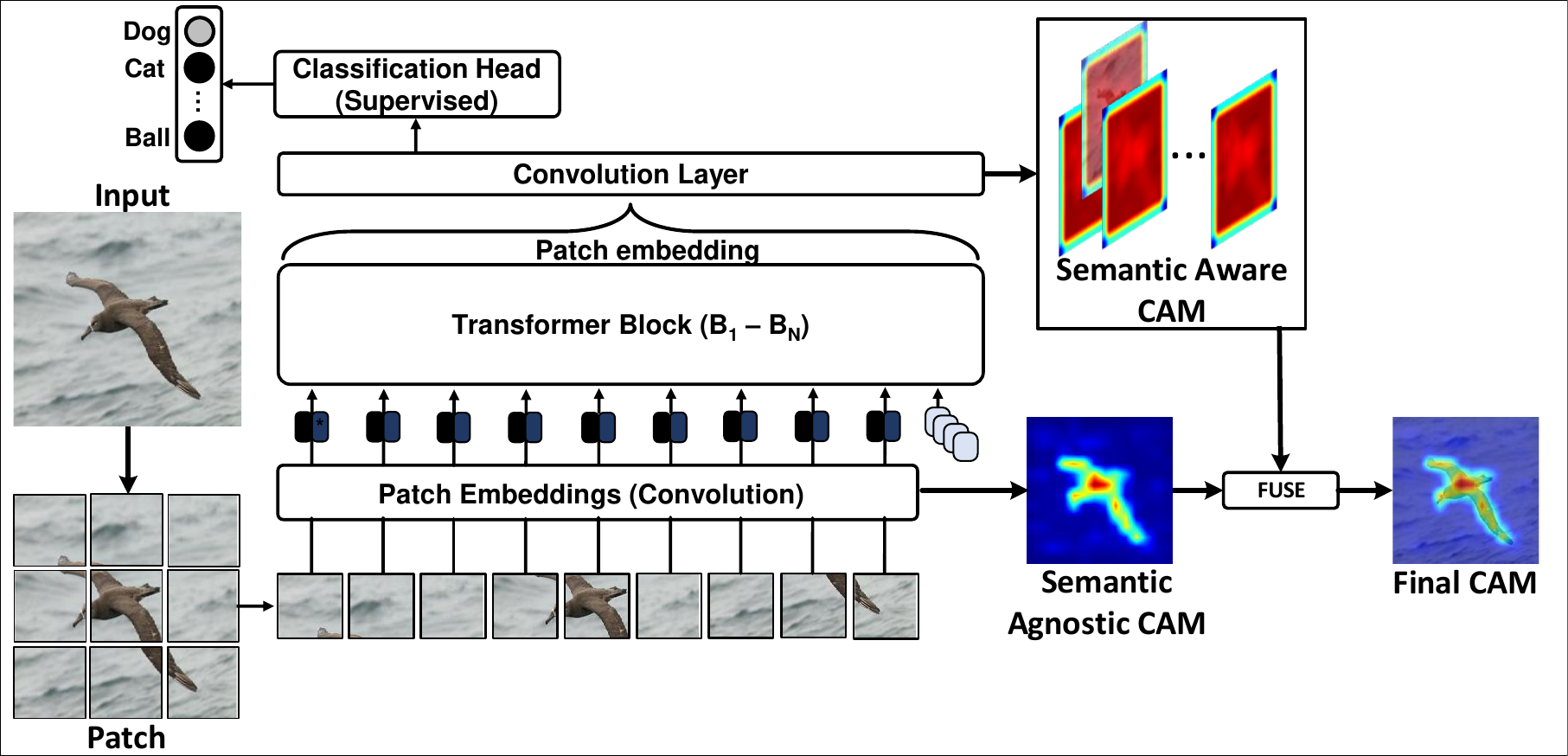}
\end{center}
 \caption{Illustration of the baseline Token Semantic Coupled Attention Map (TS-CAM) model for WSOL.}\label{fig:ts-cam}
\end{figure}
\begin{figure*}[!h]
\begin{center}
 \includegraphics[width=0.79\linewidth,trim=3 3 3 3, clip]{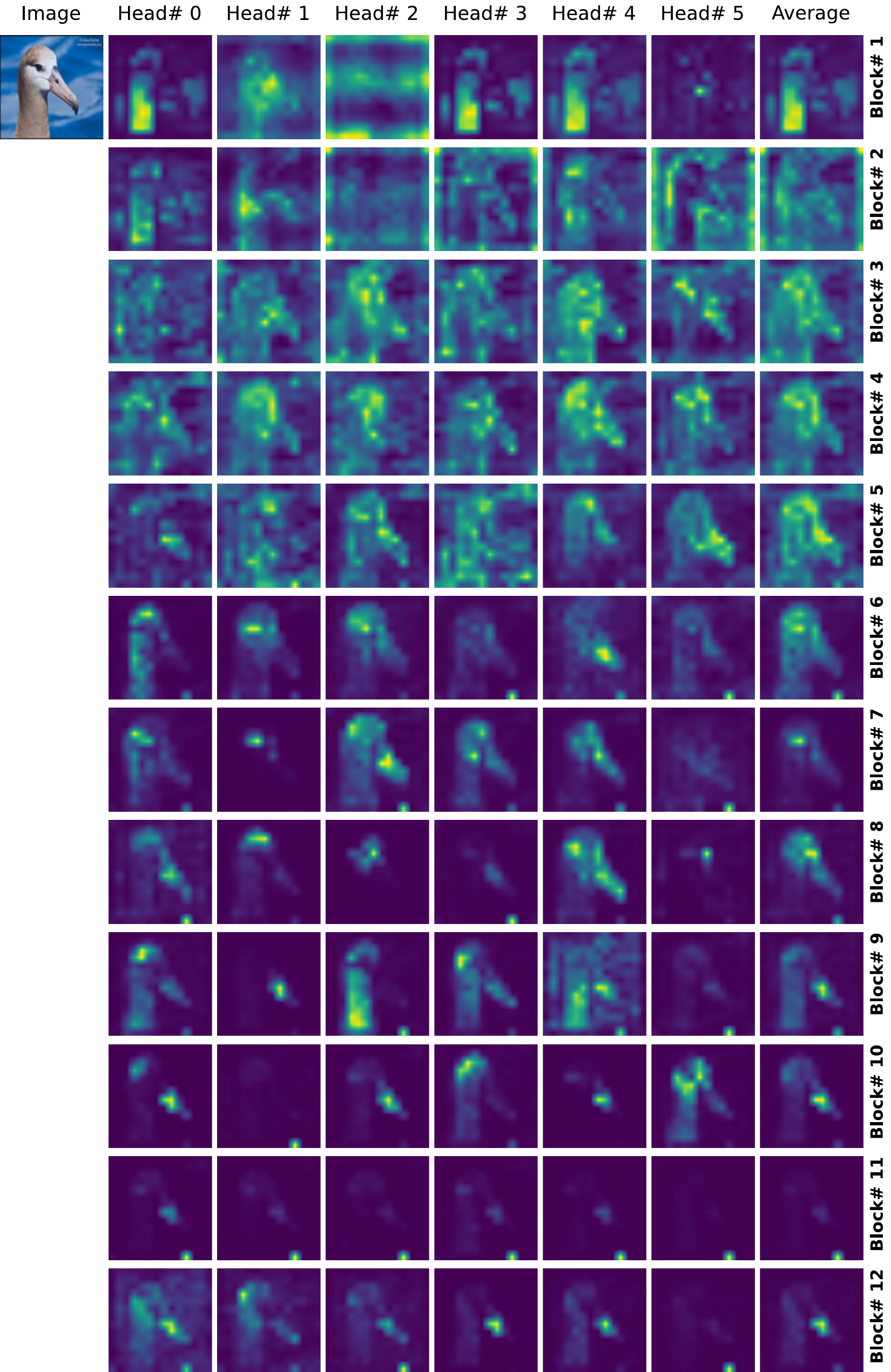}
\end{center}
\caption{Attention map of each transformer head (\cls token) learned by TS-CAM. However, different parts of the object are accumulated across the layers/blocks, and it must include semantic aware CAM to suppress noise and generate final results.}\label{app:fig_tscam_prb_ful}
\end{figure*}

By taking advantage of the attention mechanism, TS-CAM \cite{gao2021ts} is capable of capturing long-range dependency among different parts of an image. As a result, it can efficiently separate background and foreground objects. In other words, it first divides the images into a set of patches for capturing long-range dependency information and records its effects in \cls token. The attention of \cls token is then fused with the semantic aware map to produce the final attention/activation map. The flow diagram of TS-CAM is depicted in Fig.\ref{fig:ts-cam}. Lastly, a detailed visualization of the internal representation of the token TS-CAM \cls is presented in Fig.\ref{app:fig_tscam_prb_ful}. It shows that the average of all maps could potentially include noise and background regions in the final prediction.

\clearpage
\newpage


\clearpage
\newpage
{\small
\bibliographystyle{ieee_fullname}
\bibliography{egbib.bib}
}
\end{document}